\documentclass{article}

\usepackage[final]{corl_2026} 
\usepackage{amsmath}
\usepackage{amssymb}
\usepackage{graphicx}
\usepackage{booktabs}
\usepackage{tabularx}
\usepackage{multirow}
\usepackage{makecell} 
\usepackage{caption}

\title{Hardware-as-Regularization:\\
Enhancing Data Consistency in Imitation Learning \\
via an Adjustable Passive Compliant Leader Arm}

\author{
  Hideyuki Ichiwara\\
  R\&D Division\\
  SB Intuitions Corporation, 
  Japan\\
  \texttt{hideyuki.ichiwara@sbintuitions.co.jp} \\
  \And
  Yushi Kaida\\
  R\&D Division\\
  SB Intuitions Corporation, 
  Japan\\
  \texttt{yushi.kaida@sbintuitions.co.jp} \\
  \And
  Hirokazu Ishida\\
  R\&D Division\\
  SB Intuitions Corporation, 
  Japan\\
  \texttt{hirokazu.ishida@sbintuitions.co.jp} \\
}

\begin{document}
\maketitle


\begin{abstract}
    Low-cost passive leader arms enable scalable data collection for imitation learning, but their lack of restoring forces leaves redundant joints unconstrained, yielding highly inconsistent joint-space trajectories.
    We address this with a low-cost passive joint module (an adjustable mechanical clutch and a 3D-printed spiral spring) that pulls each joint toward a user-defined neutral posture.
    This passive restoring force acts as a physical prior equivalent to joint-space $L_2$ regularization (``Hardware-as-Regularization''), improving kinematic consistency with shorter completion time than an unregularized arm and preserved movement smoothness.
    Experiments show substantial variance reduction over a naive passive arm, trajectory consistency on par with an active virtual spring of the same actuator class, and substantially higher downstream success rates (e.g., 73.3\% vs.\ 36.7\% on Block Stacking); a stiffness sweep shows consistency degrading monotonically as stiffness decreases, confirming that mechanical stiffness behaves like a regularization weight.
\end{abstract}



\section{Introduction}
\label{sec:intro}

Modern robot manipulation research is increasingly driven by large-scale imitation learning and deep visuomotor policies~\citep{levine2016end,o2024open,black2025pi_}.
To acquire the necessary demonstration data at scale, the community has rapidly adopted low-cost, 3D-printable passive leader arms (e.g., ALOHA~\citep{zhao2023learning, fumobile} and GELLO~\citep{wu2024gello}).
However, this architectural simplification introduces a severe mechanical vulnerability: the complete absence of restoring forces leaves redundant degrees of freedom (DoFs) unconstrained.
Operators naturally allow their posture to drift arbitrarily in the null-space, inducing severe \textit{kinematic inconsistency}: highly multimodal joint-space trajectories that degrade the optimization landscape and sample efficiency of downstream imitation learning~\citep{belkhale2023data, feng2026demystifying}.

\begin{figure}[t]
    \centering
    \includegraphics[width=\linewidth]{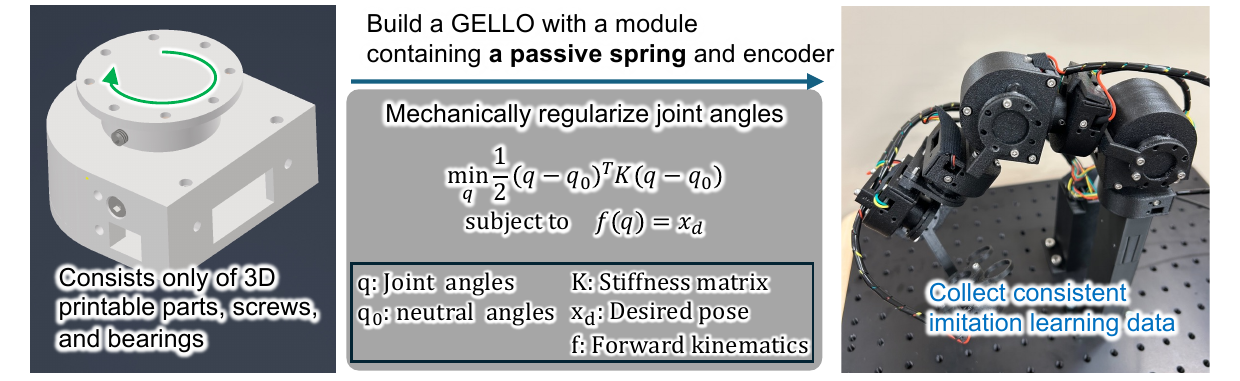}
    \caption{Overview of Hardware-as-Regularization. We introduce a purely mechanical joint module that consists only of 3D printable parts, screws, and bearings. By building a redundant teleoperation arm (e.g., GELLO) with this module containing a passive spring and an absolute encoder, we can mechanically regularize joint angles. This physical prior effectively solves the optimization problem $\min_{q} \frac{1}{2} (q - q_0)^T K (q - q_0)$ subject to the task-space constraint $f(q)=x_d$, allowing users to collect highly consistent imitation learning data.}
    \label{fig:digest}
\end{figure}

In classical robotics, kinematic redundancy is resolved via active impedance control. 
While utilizing high-end, expensive actuators can easily yield high-fidelity virtual stiffness, this fundamentally contradicts the community's core objective of scalable, democratized data collection. 
Conversely, synthesizing active virtual stiffness on modern, low-cost telemetry rigs inherently risks high-frequency instability due to control loop latency~\citep{735172}, requiring heavy artificial damping that degrades operational transparency.
To bridge the gap between the necessity for kinematic consistency and the physical limits of low-cost hardware, we introduce the paradigm of \textit{Hardware-as-Regularization} (Fig.~\ref{fig:digest}). Drawing inspiration from passive haptics~\citep{10.1115/1.2168163}, we propose a novel, purely mechanical compliant joint module featuring a Hirth coupling and a 3D-printed spiral spring. Operating in continuous time with zero latency, this spring is passive: it injects no active energy into the system.
Mathematically, it acts as a physical structural prior, applying $L_2$ regularization directly in the joint space to continuously guide the operator's posture toward a user-defined neutral point.

The main contributions of this paper are threefold: (1) a low-cost, passive (no active energy injection) joint mechanism using a 3D-printed spiral spring and an adjustable clutch; (2) the concept of \textit{Hardware-as-Regularization}, showing that physical compliance naturally resolves kinematic inconsistency in redundant teleoperation; and (3) empirical evidence that our hardware reduces joint-space variance in demonstrations, yielding downstream success rates substantially higher than a naive passive arm and comparable to an active virtual spring system of the same actuator class. Videos are available on the project page\footnote{\url{https://sbintuitions.github.io/hardware-as-regularization/}}, where the design file (STEP) will also be released.

\section{Related Work}
\label{sec:related_work}

\textbf{Low-Cost Teleoperation and Imitation Learning:}
Recent advancements heavily rely on scaling visuomotor policies via low-cost passive leader arms like ALOHA and GELLO~\citep{o2024open, wu2024gello}.
While democratizing data collection, their simplified structures introduce massive stochastic variance in the null-space, leading to state-distribution shifts that degrade Behavioral Cloning~\citep{belkhale2023data, gandhi2023eliciting}.

\textbf{Generative Policies and the Variance Bottleneck:}
State-of-the-art architectures like Diffusion Policy~\citep{chi2025diffusion} and FlowPolicy~\citep{zhang2025flowpolicy} effectively model diverse behaviors. 
However, they remain bottlenecked by the unconstrained kinematic noise of purely passive data collection, expending network capacity merely to fit variance.
This vulnerability is especially acute in joint-space representations, which are vital for control stability~\citep{feng2026demystifying}.

\textbf{Passivity and Virtual Fixtures in Redundant Control:}
Synthesizing active virtual stiffness fundamentally violates discrete-time passivity bounds~\citep{735172}, inevitably leading to unstable oscillations.
To mitigate this, researchers have explored passive haptic exoskeletons utilizing mechanical brakes or springs~\citep{1041713, 8678855}.
Building on these paradigms, we embed physical compliance directly into the hardware to mechanically regularize action divergence at the source.

\textbf{Operator Effort in Teleoperation:}
Classical models of human motor control indicate that voluntary arm movements are organized to optimize smoothness- and effort-related criteria~\citep{flash1985coordination, uno1989formation}.
This supports our modeling assumption that, in the presence of an elastic restoring force, a teleoperator tends to settle into postures minimizing physical exertion (Section~\ref{sec:consistency_problem}); we rely on behavioral proxies (completion time and smoothness) rather than direct workload measures.

\section{Hardware-as-Regularization: Concept and Design}
In this section, we formalize the kinematic inconsistency problem inherent in naive passive leader arms and demonstrate how the introduction of mechanical compliance effectively resolves this issue. We frame our approach as \textit{Hardware-as-Regularization}, where a physical spring acts as a structural prior to constrain redundant degrees of freedom (DoFs). Following this theoretical formulation, we detail the hardware design of our proposed low-cost, adjustable clutch mechanism.

\subsection{The Consistency Problem in Redundant Arms}
\label{sec:consistency_problem}

Consider a redundant leader arm with $n$ DoFs used to control a task-space pose $x \in \mathbb{R}^m$, where $n > m$. The forward kinematics can be defined as $x = f(q)$, with $q \in \mathbb{R}^n$ representing the joint configuration. The differential kinematics relating the joint velocity to the task-space velocity is given by $\Delta x = J \Delta q$, where $J \in \mathbb{R}^{m \times n}$ is the Jacobian matrix. For a desired end-effector movement $\Delta x$, the general solution for the joint displacement $\Delta q$ is expressed as $\Delta q = J^{+} \Delta x + (I - J^{+} J) z$, where $J^{+}$ is the Moore-Penrose pseudoinverse of $J$, $(I - J^{+} J)$ is the null-space projection matrix, and $z \in \mathbb{R}^n$ is an arbitrary joint velocity vector.

In naive passive arms without restoring forces, the arbitrary vector $z$ is left unconstrained. During teleoperation, gravity, friction, and human tremors continuously perturb $z$, causing unpredictable null-space drift. This results in a highly multimodal and inconsistent state-action mapping across multiple demonstrations.
To mitigate this ill-posedness without active control, we introduce a mechanical spring to the joint. Setting an arbitrary neutral posture (zero-point) $q_0 \in \mathbb{R}^n$, the passive mechanism imposes a physical potential energy on the system: $U(q) = \frac{1}{2} (q - q_0)^T K (q - q_0)$, where $K \in \mathbb{R}^{n \times n}$ is the positive definite stiffness matrix provided by the hardware.
We hypothesize that a human operator unconsciously attempts to minimize physical exertion while completing the task, in line with classical motor-control models~\citep{flash1985coordination, uno1989formation}. Therefore, operating the compliant arm can be modeled as the following constrained optimization problem:
\begin{equation}
\min_{q} \frac{1}{2} (q - q_0)^T K (q - q_0) \quad \text{subject to} \quad f(q) = x_d \label{eq:l2}
\end{equation}
where $x_d$ is the target task-space pose.
Mathematically, Equation (\ref{eq:l2}) is identical to Tikhonov regularization (L2 regularization). Because $K$ is positive definite, this physical regularization renders the objective function strongly convex. Consequently, the ill-posedness in the null-space is resolved, and the joint configuration is naturally drawn to a unique, optimal posture $q^*$. Thus, by embedding $K$ into the hardware, we provide a physical prior that absorbs operator-induced noise and automatically enforces high kinematic consistency in the collected data.

Table~\ref{tab:architecture_comparison} summarizes the fundamental architectural trade-offs between different leader arm paradigms for data collection. While naive passive arms (e.g., standard GELLO) are highly cost-effective, they suffer from low data consistency due to unconstrained null-space drift.
Conversely, augmenting these systems with active virtual springs via controllable motors can improve consistency, but this introduces significant hardware costs and conditionally stable dynamics vulnerable to control loop latency.
Our proposed mechanical approach bridges this gap; by leveraging a physical continuous-time spring, it introduces a structural prior that achieves high data consistency and passivity (no active energy is injected, in contrast to discrete-time active stiffness that is subject to passivity bounds~\citep{735172}) with zero latency, all while preserving the low cost characteristic of open-source passive systems.

\begin{table}[t]
\centering
\caption{Comparison of Leader Arm Architectures for Data Collection}
\label{tab:architecture_comparison}
\small
\begin{tabular}{@{}lccc@{}}
\toprule
\textbf{Metric} & \textbf{Naive GELLO} & \textbf{GELLO w/ v.s.} & \textbf{Proposed} \\ \midrule
\textbf{Restoring Force} & None & \makecell{Active \\ (Motor Torque)} & \makecell{\textbf{Passive} \\ \textbf{(Spiral Spring)}} \\
\textbf{Data Consistency} & \makecell{Low \\ (Null-space drift)} & High & \makecell{\textbf{High} \\ \textbf{(Physical Prior)}} \\
\textbf{Latency} & Zero & \makecell{High \\ (Control loop)} & \makecell{\textbf{Zero} \\ \textbf{(Continuous time)}} \\
\textbf{System Stability} & \makecell{Passive \\ (no energy injection)} & \makecell{Conditionally \\ Stable} & \makecell{\textbf{Passive} \\ \textbf{(no energy injection)}} \\
\textbf{Cost} & $\sim$ \$200& \makecell{$>$\$700 \\ (Current controllable motor)}& \textbf{$\sim$ \$100} \\ \bottomrule
\end{tabular}
\vspace{-0.15in}
\end{table}
\subsection{Mechanical Design}

\begin{figure}[t]
    \centering
    \includegraphics[width=\linewidth]{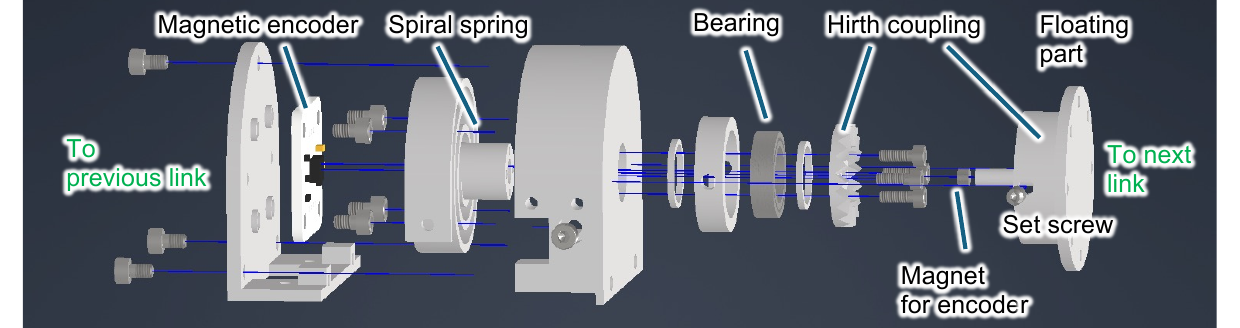}
    \caption{Exploded view of the proposed compliant joint module: a 3D-printed spiral spring for bidirectional compliance, a Hirth coupling secured via set screws for discrete neutral-position adjustment, and a coaxial magnetic encoder for joint-angle measurement.}
    \label{fig:detail}
\end{figure}

The hardware design was driven by two primary objectives: modularity to accommodate various joint architectures, and the ability to easily adjust the neutral position of the spring to adapt to diverse operational tasks.
In our design, the component designated as the ``floating part'' connects directly to the next link.
We attached a magnet to this floating part to interface with a magnetic encoder, enabling the measurement of the joint angle relative to the previous link.
All components on the opposing side (the spring assembly) are rigidly secured along the axial direction using screws.
The connection between the spring side and the floating part is achieved through axial contact via a Hirth coupling, and they are circumferentially locked into place using set screws.
Because the component fixed by the set screws incorporates a bearing, rotational movement remains unconstrained when unlocked. Consequently, operators can adjust the neutral position simply by loosening the set screws and rotating the floating part relative to the spring side. The rotational resolution of this adjustment is inherently determined by the tooth pitch of the Hirth coupling (20 degrees in our prototype).
We note that the term ``adjustable'' in this paper refers to this setup-time reconfiguration of the neutral posture $q_0$ via the clutch; the stiffness $K$ is a fabrication-time design parameter that can be set to arbitrary values by re-designing the spring geometry through the beam-theory model in Appendix~\ref{sec:appendix_spring} (see Section~\ref{sec:stiffness_sweep} for an experimental sweep).

To introduce stiffness to the axis, we employed a spiral spring.
This geometry was selected because it naturally permits bidirectional deflection around the neutral position and is highly amenable to 3D printing.
In contrast, standard torsion springs typically operate unidirectionally, requiring complex antagonistic arrangements for bidirectional compliance. 
Furthermore, since Fused Deposition Modeling yields parts weaker in the Z-axis but robust in the XY plane~\citep{10.1108/13552540210441166}, the 2D planar expansion of the spiral spring naturally aligns continuous filament lines against bending stresses, maximizing structural integrity.

Material selection was also critical for durability and performance. We fabricated the spiral spring using Polycarbonate, as Polylactic Acid, a common 3D printing filament, is highly susceptible to plastic deformation and is therefore unsuitable for compliant mechanisms. For all other structural components, we utilized Acrylonitrile Butadiene Styrene to reduce the weight of the module.


\section{Experiments}
\label{sec:experiments}

\subsection{Experimental Setup}

\begin{figure}[t]
    \centering
    \includegraphics[width=\linewidth]{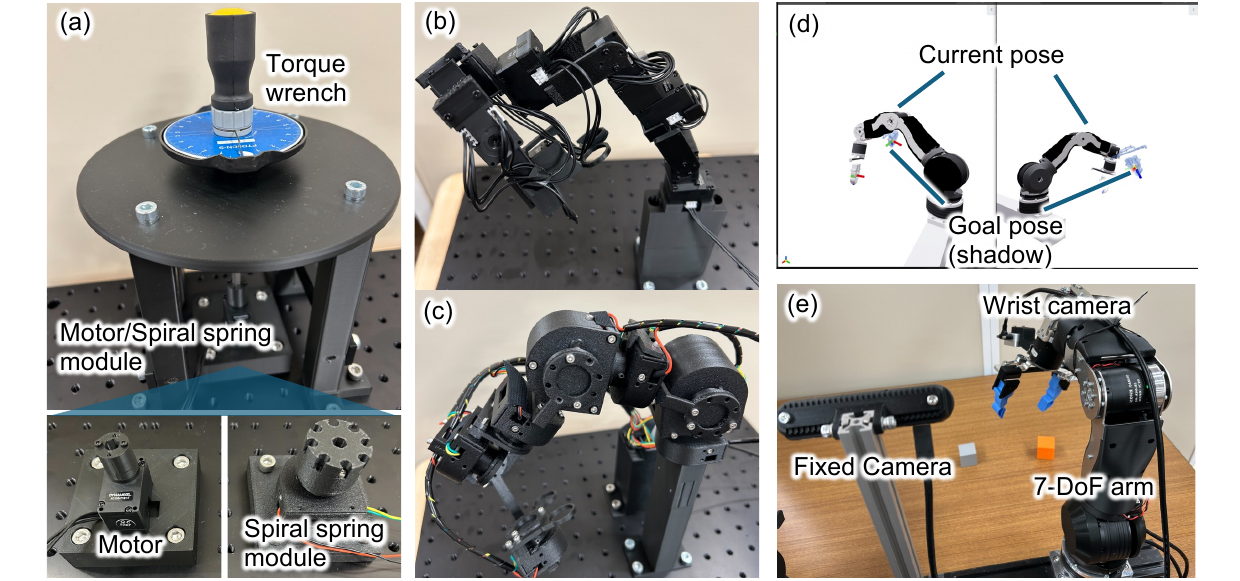}
    \caption{Overview of the experimental setups. (a) Stiffness measurement with an analogue torque wrench on the proposed module. (b, c) The leader devices: a GELLO with virtual springs and the proposed passive module. (d) The simulation interface showing the current robot pose and the target goal pose (shadow). (e) The physical workspace for imitation learning with a 7-DoF arm, a fixed camera, and a wrist camera.}
    \label{fig:experiment}
\end{figure}

We conducted three sets of experiments: a foundational assessment of the spring's mechanical stiffness, an evaluation of kinematic consistency in joint angles during teleoperation, and a performance evaluation in downstream imitation learning tasks.
To ensure a robust evaluation across different skill levels, all reaching trials were performed by a cohort of five individuals (one expert and four novices), while the imitation-learning demonstrations were collected by the expert operator.
The expert routinely collects imitation-learning demonstrations via leader--follower devices, whereas none of the four novices had any prior IL-demonstration experience with such devices. In the data consistency evaluation, the order of the three devices was randomized per participant to control for learning and order effects. All five participants gave informed consent prior to the study.
Both the leader devices and the follower OpenArm were configured as kinematically equivalent 7-DoF manipulators. All trajectory data was recorded at 30 Hz.

\subsubsection*{Stiffness Evaluation}

We first constructed a stiffness measurement system using an analogue torque wrench (Tohnichi FTD5CN-S) to evaluate the basic mechanical properties of the proposed spring module (Fig.~\ref{fig:experiment}(a)).
By applying an external force through the torque wrench, we simultaneously measured the applied torque and the resulting rotational displacement.
The torsional stiffness was then estimated using the linear least squares method. For comparison, we conducted the same measurement on a Dynamixel XC330-T181-T motor acting as a virtual spring via current control.
The proportional gain (P-gain) of the motor was tuned to approximate the stiffness of our proposed mechanical module.

\subsubsection*{Data Consistency Evaluation}

To evaluate the variance in kinematic consistency across different devices, we analyzed the joint angles produced when an operator manipulated a simulated robot arm to reach a target pose.
Specifically, we defined four spatial goal poses (up, down, left, and right relative to the front of the arm).
For each target, the operator performed 10 continuous reaching trials. We utilized leader arms constructed from three setups: Naive GELLO, GELLO with an active virtual spring, and GELLO with our proposed passive module (Figs.~\ref{fig:experiment}(b), (c)).
The teleoperation interface (Fig.~\ref{fig:experiment}(d)) was built using Viser in Python, and the follower robot was a custom desk-mounted OpenArm~\citep{openarm}.
This simulation-based targeting task was chosen because operating a real, unregularized Naive GELLO arm can result in unpredictable and potentially hazardous null-space movements.
We evaluated the trajectory consistency using Dynamic Time Warping (DTW)~\citep{sakoe1978dynamic} and assessed movement smoothness using Spectral Arc Length (SPARC)~\citep{balasubramanian2011robust, diwan2025noise}.

\subsubsection*{Imitation Learning Evaluation}

We assessed the end-to-end performance of downstream imitation learning policies using two manipulation tasks: a ``Pick \& Place'' task requiring low precision, and a ``Block Stacking'' task requiring high precision. The experimental hardware environment is shown in Fig.~\ref{fig:experiment}(e).
We employed Action Chunking with Transformers (ACT)~\citep{zhao2023learning} as our baseline policy architecture; since the module regularizes the data, not the policy, and is policy-agnostic by construction, we adopt ACT as the standard baseline of the GELLO/ALOHA lineage and leave other architectures to future work.

For training, we collected 60 demonstrations for the Pick \& Place task and 80 demonstrations for the Block Stacking task for each of the three device conditions (Naive GELLO, virtual spring GELLO, and the proposed passive module) under an identical protocol, with the expert operator manipulating the arm one-handed in all conditions, without supporting the arm with the non-operating hand---a common compensation practice for naive GELLO.
Under this unassisted condition, naive GELLO collection proved hazardous: unplanned self-weight droop of null-space joints repeatedly caused collisions or extension-limit locking, forcing emergency stops in $\sim$12.5\% of attempts ($\sim$20 out of $\sim$160); aborted trials were excluded and recollected.
This quantifies the operational risk of the unregularized arm and motivates our simulation-based consistency evaluation.
The IL collection order was fixed with the naive condition last, so practice effects favor the naive condition and the gaps reported below are conservative.
The trained policies were evaluated over 30 autonomous trials per task; alongside success rates, we analyzed dataset quality using the same consistency metrics.

\subsection{Analysis of Stiffness}
\label{sec:stiffness_analysis}

\begin{figure}[t]
    \centering
    \includegraphics[width=\linewidth]{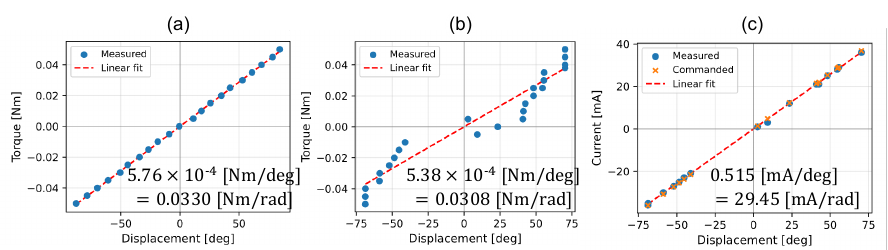}
    \caption{Results of the stiffness evaluation. (a) Torque versus displacement for the proposed passive spiral spring module (linear fit: 0.0330 Nm/rad). (b) Torque versus displacement for the active virtual spring, showing intermittent mechanical dead zones. (c) Commanded and measured current versus displacement for the virtual spring (linear electrical response: 29.45 mA/rad).}
    \label{fig:stiffness}
\end{figure}

Assuming both ends of the spring are rigidly mounted, the theoretical torsional stiffness $K$ based on beam theory is approximately $0.025$~Nm/rad (equivalent to $25$~N$\cdot$mm/rad)~\citep{ahmed2014synthesis}.
Detailed formulations and parameter derivations are provided in Appendix~\ref{sec:appendix_spring}. Because our design embeds the inner end rigidly into the central axis and fixes the outer end completely, we use this $0.025$~Nm/rad value as our baseline model.
Figure~\ref{fig:stiffness}(a) illustrates the measured torque against angular displacement for the proposed module. The empirical results confirm a highly linear spring characteristic, with a measured stiffness of $0.0330$~Nm/rad. This slight increase from the theoretical prediction is primarily attributed to 3D printing manufacturing tolerances and internal friction. However, this minor discrepancy poses no practical issue for regularization. Rather than achieving absolute hardware identicality across different setups, our primary goal is ensuring intra-dataset consistency for a given operator-device pair. Minor stiffness variations are naturally absorbed by human adaptability, maintaining highly consistent trajectory data without requiring rigorous mechanical tuning. Furthermore, the proposed module demonstrated high mechanical reliability, exhibiting minimal stiffness variation across identically printed modules ($0.0322$--$0.0335$~Nm/rad) and negligible degradation after a 1,000-cycle continuous reciprocating test.

In contrast, Figure~\ref{fig:stiffness}(b) shows the torque-displacement curve for the motor acting as a virtual spring. Intermittent mechanical dead zones are clearly observable across the displacement range. Although the current output remains strictly proportional to the displacement (Fig.~\ref{fig:stiffness}(c)), this dead zone is likely caused by inherent gear friction and backlash within the actuator.
This discontinuity suggests that active virtual springs introduce a jerky, non-transparent operational feel, precisely where a continuous-time, zero-latency physical spring excels.
We deliberately implemented the active baseline as a pure proportional virtual spring on the same actuator class, with the P-gain tuned to approximate our mechanical stiffness, so as to contrast two \emph{realizations} of the same regularization.
An added damping term cannot remove the measured dead zone: Fig.~\ref{fig:stiffness}(b) is a quasi-static measurement, where viscous damping ($\propto$ velocity) vanishes.
Empirically, even our $K/2$ springs significantly outperform this nominally full-stiffness baseline in consistency (Section~\ref{sec:stiffness_sweep}).

Finally, a free-decay test (releasing the arm without hand contact; Appendix~\ref{sec:appendix_freedecay}) shows underdamped ringing in the excited joints ($\zeta \approx 0.09$--$0.16$, settling in $\sim$1--5~s) and a small overdamped response in the barely excited base swivel. The module is thus passive rather than oscillation-free; during data collection, hand contact supplies damping, and completion time and smoothness are unaffected (Section~\ref{sec:kinematic_consistency}).

\subsection{Analysis of Kinematic Consistency}
\label{sec:kinematic_consistency}

\begin{figure}[t]
    \centering    
    \begin{minipage}[c]{0.3\linewidth}
        \centering
        \includegraphics[width=\linewidth]{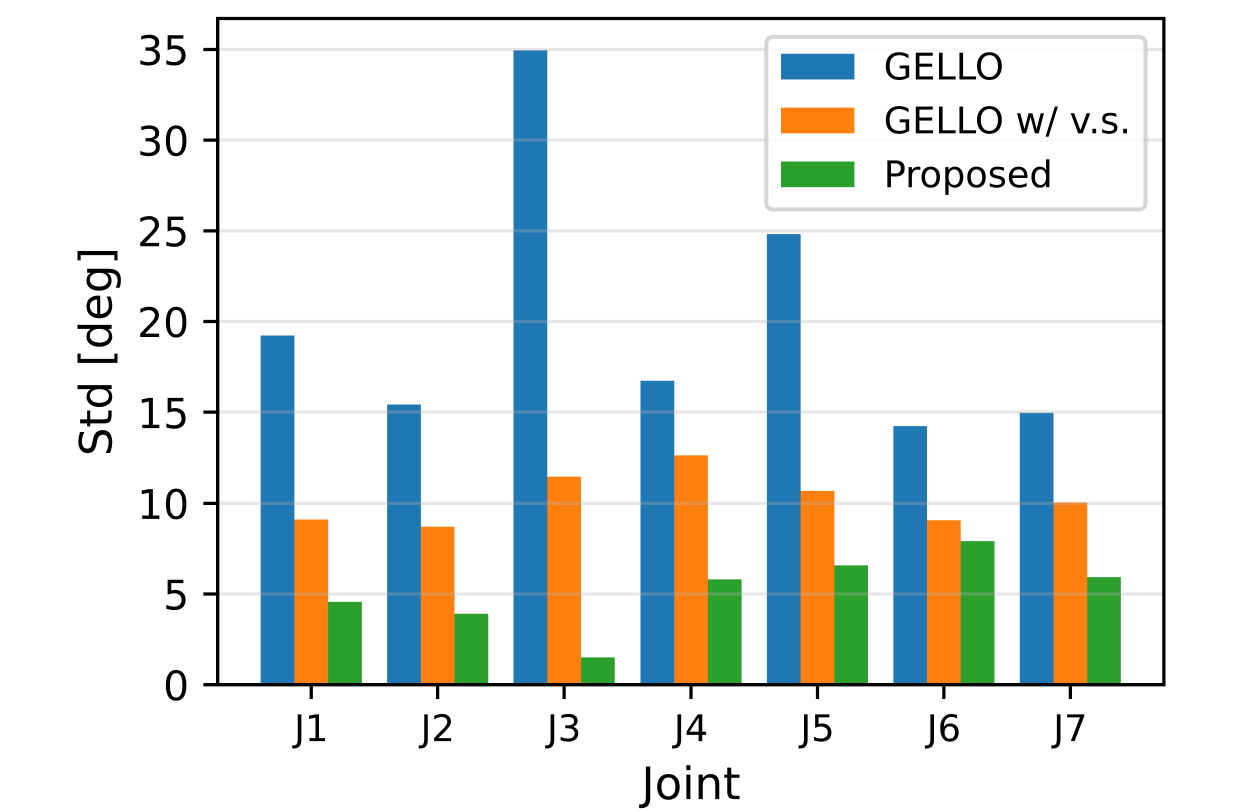}
        \caption{Comparison of joint angle standard deviations across different leader architectures.}
        \label{fig:joint_std}
    \end{minipage}
    \hfill
    \begin{minipage}[c]{0.66\linewidth}
        \centering
        \footnotesize
        
        \setlength{\tabcolsep}{3pt} 
        
        \begin{tabular}{lrrr}
            \toprule
            Metric & Naive GELLO& GELLO w/ v.s. & Proposed \\ 
            \midrule
            Joint std [deg] & 20.06 $\pm$ 5.13 & 10.25 $\pm$ 1.46 & \textbf{5.17 $\pm$ 0.36} \\
            Pairwise DTW [deg]  & 37.19 $\pm$ 18.64 & 22.01 $\pm$ 7.39 & \textbf{11.82 $\pm$ 3.91} \\
            SPARC [-]  & -3.26 $\pm$ 1.42 & -4.23 $\pm$ 3.00 & -2.61 $\pm$ 0.94 \\
            \bottomrule
        \end{tabular}
        \captionof{table}{Quantitative metrics comparing joint angle variance, DTW, and trajectory smoothness (SPARC) across devices and operators. (Note: For SPARC, values closer to zero indicate smoother trajectories)}
        \label{tab:consistency}
    \end{minipage}
    
\end{figure}

Figure~\ref{fig:joint_std} presents the standard deviation of joint angles for each device.
The Naive GELLO exhibits massive variance, particularly in the joints associated with the manipulator's null-space (e.g., J2 and J4).
While original literature often focuses on suppressing movement strictly in the direction of gravity, these results clearly indicate that regularizing other redundant joints is equally critical.
Both the GELLO with virtual springs and our proposed method significantly reduce this variance, demonstrating that physical/virtual regularization effectively mitigates drift.

\begin{figure}[t]
    \centering
    \includegraphics[width=\linewidth]{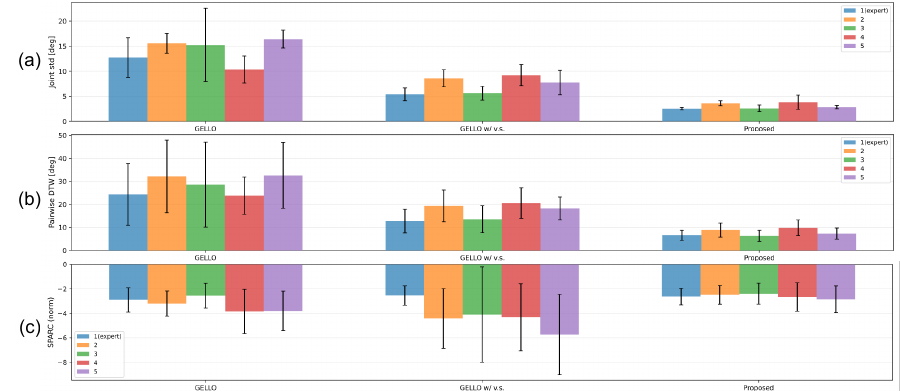}
    \caption{Comparison of individual operator performance across three kinematic consistency metrics: (a) joint standard deviation, (b) Pairwise DTW, and (c) trajectory smoothness (SPARC). Note that User 1 is an expert, while Users 2–5 are novices. The proposed module consistently minimizes null-space drift (a, b) for each user regardless of skill level, while preserving natural smoothness (c).}
    \label{fig:operator_variance}
\end{figure}

As detailed in Table~\ref{tab:consistency}, the proposed method achieved the lowest joint variance and Pairwise DTW, confirming that the mechanical prior enforces highly consistent null-space postures.
Regarding trajectory smoothness, Table~\ref{tab:consistency} shows the proposed method yields a highly stable SPARC score ($-2.61 \pm 0.94$). While the naive GELLO maintained moderate smoothness due to natural human damping, the active virtual spring exhibited a lower mean with substantially larger variance ($-4.23 \pm 3.00$). Although this variance precludes claims of strict statistical superiority, it supports our stiffness analysis: inherent gear backlash and friction in the active motor occasionally introduce unpredictable, jerky operational feels. In contrast, our continuous-time physical spring provides seamless compliance, preventing this degradation. Ultimately, our purely mechanical system matches, and in trajectory smoothness, structurally improves upon the regularization quality of an active setup.
As Fig.~\ref{fig:operator_variance} shows, the proposed method consistently minimized intra-operator variance for each of the five operators, regardless of experience: despite distinct individual kinematic preferences, the mechanical prior suppresses null-space drift for all users, enabling consistent data collection without extensive training.

Regarding operator load, we did not measure cognitive load directly. As behavioral proxies, completion time in the expert consistency trials was shorter with the proposed device than with the naive arm (4.42 s vs.\ 6.56 s on average; Welch's $t$-test, $p=7.6\times10^{-3}$), with smoothness maintained (Table~\ref{tab:consistency}). Together with the uninstructed, null-space-localized drift suppression (J2/J4, Fig.~\ref{fig:joint_std}), these observations are consistent with the exertion-minimization hypothesis.

\subsection{Effect of Stiffness as a Regularization Weight}
\label{sec:stiffness_sweep}

To exercise the interpretation of $K$ as a regularization weight, we replaced all seven joints' springs with thickness-scaled counterparts of approximately half the design stiffness ($K/2$), following the beam-theory model of Appendix~\ref{sec:appendix_spring} (a torque-wrench check on a sampled spring confirmed the halving), and repeated the simulation reaching protocol with the same expert operator (same-operator comparison).
Consistency improves monotonically as $K$ increases ($K=0 \rightarrow K/2 \rightarrow K$: joint std 12.72 $\rightarrow$ 3.85 $\rightarrow$ 2.55 deg; pairwise DTW 24.20 $\rightarrow$ 10.17 $\rightarrow$ 6.86 deg), with all adjacent DTW gaps significant (Welch's $t$-test; e.g., $K/2$ vs.\ $K$: $p=3.5\times10^{-6}$). This dose--response supports the Tikhonov interpretation of $K$ as a regularization weight; conversely, excessively large $K$ would increase operator effort. The virtual spring, tuned to approximate the full stiffness $K$, achieves only an intermediate consistency (joint std 5.39 deg, DTW 13.07 deg), significantly worse than even our half-stiffness springs (DTW, $p=3.6\times10^{-3}$), consistent with the backlash/deadzone analysis of Section~\ref{sec:stiffness_analysis}.

\subsection{Imitation Learning Performance}

\begin{table}[t]
    \centering
    \caption{Comparison of collected-data quality and downstream ACT success rates (95\% Wilson score intervals; $n=30$ rollouts per condition). Naive GELLO trails the proposed module significantly on both tasks (two-sided Fisher exact test), while proposed vs.\ virtual spring is not significant on either task (see text).}
    \label{tab:il_performance}
    \small
    \setlength{\tabcolsep}{3pt}
    
    \begin{tabular}{@{}llrrr@{}} 
        \toprule
        Task & Metric & Naive GELLO & GELLO w/ v.s. & Proposed \\ 
        \midrule
        
        \textbf{Pick \& Place} 
        & Pairwise DTW [deg] (Data) & 23.23 $\pm$ 11.01 & 11.71 $\pm$ 5.51 & 8.32 $\pm$ 3.14 \\
        & SPARC [-] (Data) & -2.87 $\pm$ 0.23 & -3.23 $\pm$ 0.23 & -2.97 $\pm$ 0.25 \\
        \addlinespace 
        & Success Rate (95\% CI) [\%] & 53.3 (36.1,\, 69.8) & 96.7 (83.4,\, 99.4) & 96.7 (83.4,\, 99.4) \\
        
        \midrule
        
        \textbf{Stack} 
        & Pairwise DTW [deg] (Data) & 16.35 $\pm$ 6.94 & 10.23 $\pm$ 3.99 & 8.24 $\pm$ 2.99 \\
        & SPARC [-] (Data) & -2.87 $\pm$ 0.21 & -3.89 $\pm$ 0.30 & -3.19 $\pm$ 0.33 \\
        \addlinespace
        & Success Rate (95\% CI) [\%] & 36.7 (21.9,\, 54.5) & 66.7 (48.8,\, 80.7) & 73.3 (55.6,\, 85.8) \\
        
        \bottomrule
    \end{tabular}
\end{table}

Table~\ref{tab:il_performance} summarizes dataset quality and ACT policy success rates.
The Naive GELLO baseline lags far behind both regularized setups: its joint-space DTW is 2.0--2.9$\times$ worse than that of the proposed module (23.23 vs.\ 8.32 deg on Pick \& Place and 16.35 vs.\ 8.24 deg on Stack; Welch's $t$-test, $p<5\times10^{-15}$ for both), and its ACT success rates are lower by large, statistically significant margins (53.3\% vs.\ 96.7\% on Pick \& Place, and 36.7\% vs.\ 73.3\% on Stack; two-sided Fisher exact test, $p=1.7\times10^{-4}$ and $p=8.9\times10^{-3}$, respectively).
Notably, the naive demonstrations remained smooth: their SPARC scores are as good as or better than those of the regularized setups, consistent with natural human damping. This indicates that joint-space multimodality, not smoothness, drives the $\sim$2$\times$ drop in success rate. This completes the causal chain of our claim: restoring forces reduce demonstration variance (Section~\ref{sec:kinematic_consistency}), and reduced variance improves imitation learning.

Comparing the two regularized setups, the proposed module and the virtual spring baseline showed comparable trajectory consistency (Pairwise DTW) with overlapping standard deviations, and their success rates do not differ significantly on either task; the identical 96.7\% on Pick \& Place reflects its low precision demand, yet the task discriminates clearly once the naive condition is included.
A difference in trajectory smoothness (SPARC) emerged depending on the required precision. In Pick \& Place, which predominantly involves macro-movements, both devices yielded equivalent smoothness. Conversely, in Block Stacking, which demands continuous fine positional adjustments, the proposed module exhibited significantly better SPARC scores than the virtual spring ($-3.19 \pm 0.33$ vs.\ $-3.89 \pm 0.30$; Welch's $t$-test, $p=1.7\times10^{-29}$).
We attribute this to the mechanical dead zone of the active motor identified in our stiffness analysis (Section~\ref{sec:stiffness_analysis}): during delicate manipulations, gear backlash and friction likely introduce high-frequency micro-stutters, whereas our continuous-time physical spring provides zero-latency, seamless compliance, preserving smoothness during fine posture corrections.

\section{Discussion and Conclusion}
\label{sec:discussion}
We introduced \textit{Hardware-as-Regularization}: a low-cost, purely mechanical 3D-printed joint module that resolves kinematic inconsistency in redundant passive leader arms. Its passive, zero-latency physical prior reduces joint-angle variance, improves downstream imitation learning over the naive passive arm by a significant margin, and matches an active virtual spring system of the same actuator class in success rate while improving smoothness.
Across skill levels, the prior robustly suppresses null-space drift, enforcing tight \textit{intra-operator} consistency.

\textbf{Limitations and future work.}
Our active baseline is a proportional virtual spring on the same actuator class; comparisons against PD controllers with backlash compensation are future work.
The stiffness sweep used a single expert; downstream learning used only ACT (the module regularizes the data, not the policy); the IL collection order was fixed, albeit conservatively.
Neither the exertion-minimization hypothesis nor cognitive load was measured directly; our claims rest on behavioral proxies and the observed dose--response, with standardized workload instruments (e.g., NASA-TLX~\citep{hart1988development}) as future work.
Neutral-posture reconfiguration takes $\sim$2--3 minutes across the seven joints and spring replacement $\sim$30 minutes excluding printing (Appendix~\ref{sec:appendix_spring}); these are setup-time operations, albeit slower than the near-instant software alternative.
Restoring forces may induce fatigue on workspace-spanning dynamic tasks, though most manipulation is localized; spring fatigue is offset by open-source repairability. Larger manipulators demand stiffer, bulkier springs; easier packaging and metamaterials are future work.

\clearpage
\acknowledgments{We thank all study participants for their time and contribution.
This work was fully supported by SB Intuitions Corp.}


\bibliography{example}  


\clearpage
\appendix

\section{Hardware Specifications and Fabrication Details}
\label{sec:appendix_hw}
This section provides detailed specifications of the robotic systems, the theoretical and empirical mechanical properties of the proposed passive module, and the fabrication guidelines necessary to reproduce our hardware setup.

\subsection{Robot Configuration and Teleoperation Mapping}

\begin{figure}[htbp]
    \begin{minipage}[c]{0.55\textwidth}
        \centering
        \captionof{table}{Arm joint specifications.}
        \label{tab:arm-kinematics}
        \begin{tabular}{@{}ccccccc@{}}
            \toprule
            Joint & $x$ & $y$ & $z$ & Axis & Range \\
                  & [mm] & [mm] & [mm] &      & [deg] \\
            \midrule
            J1 & 0.0 & 0.0 & 62.5 & Z & $[-50, 230]$ \\
            J2 & -30.1 & 0.0 & 60.0 & -X & $[-10, 190]$ \\
            J3 & 30.1 & 0.0 & 66.2 & Z & $[-90, 90]$ \\
            J4 & -0.0 & 31.5 & 153.8 & Y & $[0, 140]$ \\
            J5 & 0.0 & -31.5 & 95.5 & Z & $[-90, 90]$ \\
            J6 & 37.5 & 0.0 & 120.5 & X & $[-45, 45]$ \\
            J7 & -37.5 & 0.0 & 0.0 & Y & $[-90, 90]$ \\
            \bottomrule
        \end{tabular}
    \end{minipage}\hfill
    \begin{minipage}[c]{0.4\textwidth}
        \centering
        \includegraphics[width=\linewidth]{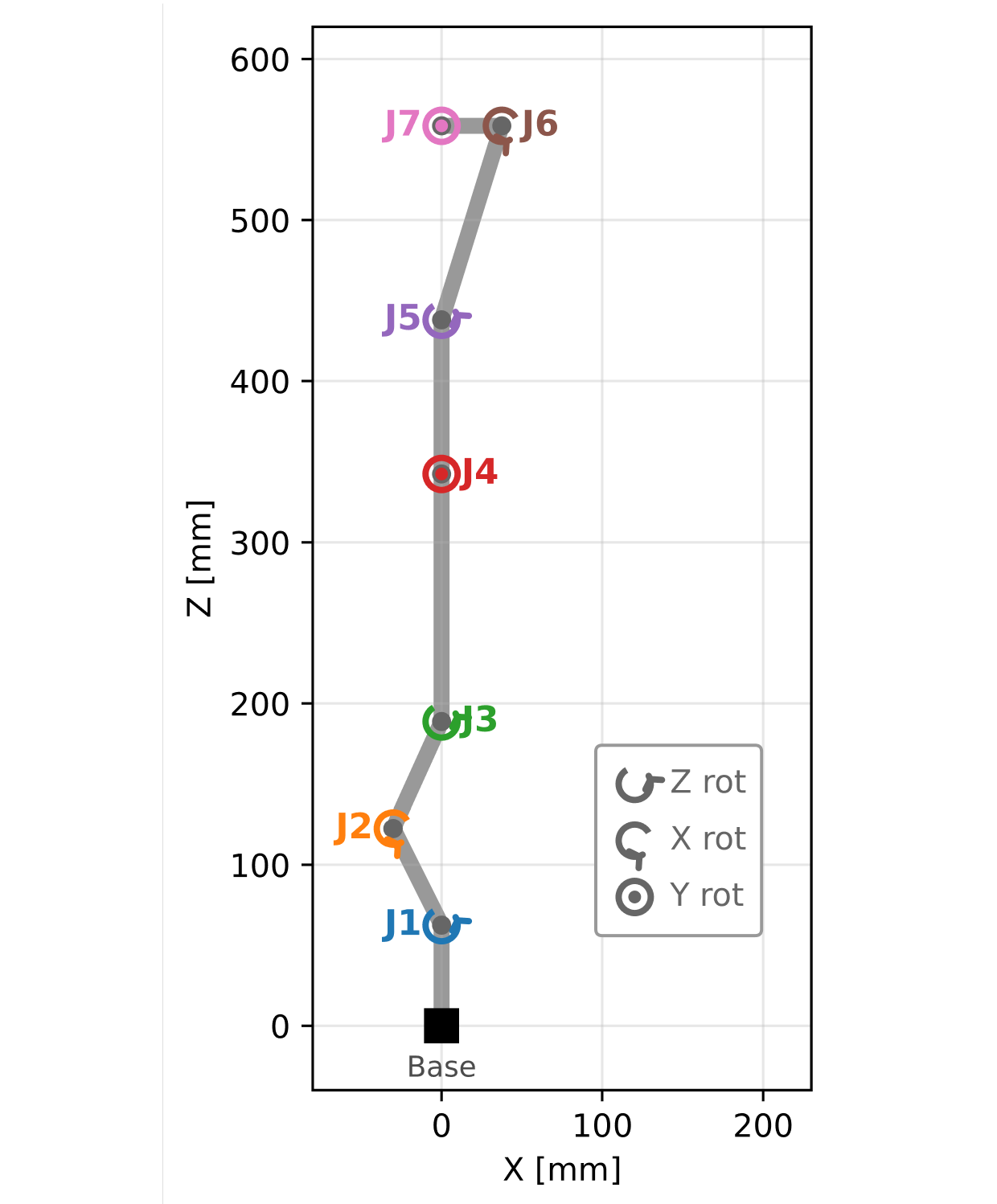}
        \caption{Joint configuration and axes.}
        \label{fig:config}
    \end{minipage}
\end{figure}

The joint configuration and axes for the robotic arms are detailed below. Table~\ref{tab:arm-kinematics} outlines the specific dimensional and rotational parameters, while Figure~\ref{fig:config} visually illustrates the spatial arrangement of each joint. Although the leader devices are scaled to approximately 40\% of the physical follower robot, the joint angles are mapped directly (1:1) from the leader to the follower arm during teleoperation, maintaining a consistent joint-space control loop.

\subsection{Leader Device Configurations}

In our experiments, we constructed leader devices based on the Naive GELLO and our proposed module. Both devices were scaled to approximately 40\% of the physical follower robot. Crucially, the hardware configuration and parameter settings of the leader devices were kept uniform and identical across all operators, independent of their individual experience or skill levels.

For the Naive GELLO setup, Dynamixel XC330-T181-T actuators were used for all joints except J2. Because J2 bears a higher load, we used a Dynamixel XC330-T288-T, which maintains the same form factor but provides higher output torque.

For the leader device equipped with the proposed module, all joints except J2 and J4 share the same design parameters used in our stiffness evaluation ($b = 8\text{ mm}$, $t = 1.5\text{ mm}$, $N = 3$). To adequately support the joints driven heavily by gravity, we modified the parameters of J2 and J4 to significantly increase their mechanical stiffness:
\begin{itemize}
    \item \textbf{J2:} $b = 30\text{ mm}$, $t = 2.5\text{ mm}$, $p = 5.9\text{ mm}$, $N = 3$ (Theoretical stiffness: $0.344\text{ Nm/rad}$)
    \item \textbf{J4:} $b = 25\text{ mm}$, $t = 2.0\text{ mm}$, $p = 5.5\text{ mm}$, $N = 3$ (Theoretical stiffness: $0.153\text{ Nm/rad}$)
\end{itemize}
For the active baseline (GELLO w/ virtual spring), the proportional gain ($K_p$) was set to $50$ for all joints except J2 and J4, which were set to $100$.

\subsection{Mechanical Properties of the Spiral Spring}
\label{sec:appendix_spring}

Assuming both ends of the spring are rigidly mounted, the theoretical torsional stiffness $K$ is defined based on beam theory as follows~\citep{ahmed2014synthesis}:
\begin{equation}
    K = \frac{E I}{L} = \frac{E b t^3}{12 L} \label{eq:stiffness}
\end{equation}
where $E$ is the flexural modulus of the material, $I$ is the area moment of inertia, $b$ is the width, $t$ is the thickness, and $L$ is the unrolled length of the spring.
Equation~(\ref{eq:stiffness}) expresses the rate per radian; the handbook form $\pi E b t^3 \theta / (6L)$ in~\citep{ahmed2014synthesis} gives the identical rate with the deflection $\theta$ measured in revolutions.
The unrolled length $L$ of an Archimedean spiral spring can be approximated using the inner radius $r_{in}$, the outer radius $r_{out}$ (where $r_{out} = r_{in} + p N$, with $p$ as the pitch and $N$ as the active number of coils), and the average radius $r_{avg} = (r_{in} + r_{out})/2$:
\begin{equation}
    L \approx 2\pi r_{avg} N \label{eq:length}
\end{equation}
Applying the parameters of our design ($E = 2300$~MPa, $b = 8$~mm, $t = 1.5$~mm, $r_{in} = 5$~mm, $p = 4$~mm, $N = 3$), the outer radius becomes $r_{out} = 17$~mm, and the average radius is $r_{avg} = 11$~mm. This yields an area moment of inertia $I = 2.25$~mm$^4$ and an unrolled length $L \approx 207$~mm, resulting in a theoretical stiffness of approximately $K \approx 0.025$~Nm/rad (equivalent to $25$~N$\cdot$mm/rad). This theoretical value represents an upper bound under ideal rigid conditions. In practical applications, if the outer end is pin-supported (hinged), the constraining moment is released, and stiffness decreases (typically to 0.8 times the pure bending value)~\citep{ahmed2014synthesis}. Because our design embeds the inner end rigidly into the central axis and fixes the outer end completely, we use the value derived from Equation (\ref{eq:stiffness}) as our baseline model.

The empirical results confirm a highly linear spring characteristic; however, the measured stiffness (0.0330~Nm/rad) was higher than the theoretical prediction. This discrepancy is attributed to manufacturing tolerances in thickness, a slightly reduced functional length, and internal friction. Typical FDM 3D printers exhibit dimensional errors of 0.2 to 0.5~mm. Because stiffness is proportional to the cube of the thickness, an error of just +0.2~mm increases the stiffness by a factor of 1.45. Additionally, the design embeds the spiral partially into the shaft, reducing the effective length by approximately 10~mm and increasing theoretical stiffness by another 5\%.

Crucially, while there is a discrepancy from the theoretical model, the FDM 3D printing process yields highly reproducible results under identical settings. We fabricated four identical spiral springs and measured their stiffness to evaluate manufacturing variance. The results showed minimal variation across the four samples, ranging narrowly from 0.0322 to 0.0335~Nm/rad. This high intra-batch consistency ensures that while the absolute stiffness may deviate from theory, the physical properties remain predictable and reliable for data collection across multiple modules.

For the stiffness sweep experiment reported in Section~\ref{sec:stiffness_sweep}, we manufactured thickness-scaled counterparts of all seven joints' springs whose design stiffness was approximately halved ($K/2$) according to Equation~(\ref{eq:stiffness}). A torque-wrench check on a sampled spring confirmed that the measured stiffness was approximately halved ($0.016$~Nm/rad), which we judged sufficient for the purposes of the sweep. Replacing the springs on all seven joints takes approximately 30~minutes excluding printing time; owing to the modular design, the change is confined to the spring itself (only a spacer accommodates its slightly different radius) while all other module components are reused.

\subsection{Free-Decay Test}
\label{sec:appendix_freedecay}

To characterize the dynamic response of our passive module, we performed a free-decay test in which the end-effector was lifted and released without hand contact (10 trials). Encoder readings of each joint were recorded during the decay. As summarized in Table~\ref{tab:free_decay}, the joints excited by this perturbation (J2--J7) exhibited underdamped ringing with damping ratios $\zeta \approx 0.09$--$0.16$ and settling times of approximately 1--5~s. The base swivel joint J1, which was barely excited by this perturbation, showed only a small overdamped response. During data collection, continuous hand contact supplies additional damping; consistently, we observed no degradation in completion time or movement smoothness (Section~\ref{sec:kinematic_consistency}).

\begin{table}[htbp]
    \centering
    \caption{Free-decay test summary (10 trials). $\zeta$: damping ratio, $\omega_n$: natural frequency.}
    \label{tab:free_decay}
    \small
    \begin{tabular}{@{}lccc@{}}
        \toprule
        Joint & Settling [s] & $\zeta$ & $\omega_n$ [Hz] \\ \midrule
        J1 & $4.92 \pm 0.12$ & overdamped & n/a \\
        J2 & $1.91 \pm 0.18$ & $0.15 \pm 0.01$ & $1.61 \pm 0.09$ \\
        J3 & $5.05 \pm 0.11$ & $0.09 \pm 0.03$ & $3.24 \pm 0.43$ \\
        J4 & $2.23 \pm 0.15$ & $0.09 \pm 0.02$ & $2.59 \pm 0.20$ \\
        J5 & $1.90 \pm 1.08$ & $0.13 \pm 0.11$ & $2.13 \pm 0.53$ \\
        J6 & $3.49 \pm 1.44$ & $0.10 \pm 0.07$ & $3.19 \pm 0.97$ \\
        J7 & $1.04 \pm 0.02$ & $0.16 \pm 0.02$ & $2.66 \pm 0.89$ \\ \bottomrule
    \end{tabular}
\end{table}

\subsection{3D Printing Settings}
All components were fabricated using a Bambu Lab P1S 3D printer, and the slicing was handled by Bambu Studio. We primarily used the default profiles automatically provided by the slicing software. However, to optimize the printing of the Polycarbonate (PC) spiral springs, we applied two specific modifications:
\begin{enumerate}
    \item \textbf{Wall Loops:} Increased to 6 to enhance the structural stiffness and durability of the springs.
    \item \textbf{Brim Width:} Set to $10\text{ mm}$ to prevent part warping and ensure bed adhesion during the print process.
\end{enumerate}

\subsection{Bill of Materials (BOM)}
Table~\ref{tab:bom} outlines the cost breakdown for constructing one leader arm equipped with our proposed modules. The AS5600 magnetic encoders at each joint communicate via the I2C protocol. Since these sensors share identical default hardware addresses, a TCA9548A I2C multiplexer is integrated to aggregate and route the individual joint signals seamlessly to a single Seeed Studio XIAO RP2040 microcontroller for high-frequency angle processing. Note that while the filament costs are listed per spool, a single spool provides sufficient material to print the entire arm structure.

\begin{table}[htbp]
    \centering
    \caption{Bill of Materials for the Proposed Leader Arm}
    \label{tab:bom}
    \begin{tabular}{@{}llc@{}}
        \toprule
        \textbf{Component} & \textbf{Quantity} & \textbf{Estimated Cost (USD)} \\ \midrule
        Polycarbonate (PC) Filament & 1 spool & \$45 \\
        ABS Filament & 1 spool & \$15 \\
        AS5600 Magnetic Encoder & 8 units & \$15 \\
        TCA9548A I2C Multiplexer & 1 unit & \$2 \\
        Seeed Studio XIAO RP2040 & 1 unit & \$8 \\
        6700ZZ Bearing & 8 units & \$10 \\
        M2/M2.5 Screws and Nuts & 1 set & $\sim$\$10 \\ \midrule
        \textbf{Total} & & \textbf{$\sim$\$105} \\ \bottomrule
    \end{tabular}
\end{table}

\section{Experimental Setups and Learning Details}
\label{sec:appendix_exp}
This section outlines the specific task environments, success criteria, and neural network configurations utilized to evaluate the efficacy of the collected imitation learning data.

\subsection{Task Definitions and Evaluation Criteria}

\textbf{1. Data Consistency Evaluation (Reaching Task):}

To evaluate the variance in the null-space posture, operators were tasked with manipulating the simulated robotic arm to reach four predefined spatial goal poses. To enforce a wide spatial range of motion, the goal poses were sequenced to move Right $\rightarrow$ Left $\rightarrow$ Up $\rightarrow$ Down from the perspective of the base position. The target poses represented as $(x, y, z, q_x, q_y, q_z, q_w)$ are as follows:
\begin{itemize}
    \item Right: $(0.3, -0.2, 0.2, 0.030, 0.970, -0.171, 0.171)$
    \item Left: $(0.3, 0.2, 0.2, -0.209, 0.929, 0.280, 0.121)$
    \item Up: $(0.3, 0.0, 0.5, -0.021, 0.762, -0.021, 0.647)$
    \item Down: $(0.3, 0.0, 0.0, -0.087, 0.992, 0.008, 0.087)$
\end{itemize}
A reaching trial was determined to be successfully completed (goal reached) when the follower robot's end-effector pose was within a position threshold of 50~mm and an orientation threshold of 15~degrees relative to the target goal pose.

\textbf{2. Imitation Learning Evaluation (Pick \& Place and Stack):}

We evaluated the system on two manipulation tasks: Pick \& Place and Block Stacking.
For the Pick \& Place task, as shown in Fig.~\ref{fig:exsetup}, the target object is a replica banana, and the placement destination is a plate. The operational workspace is approximately $450\text{ mm} \times 300\text{ mm}$. The task requires grasping the banana located in the right half of the workspace and placing it onto the plate in the left half. The success criterion is satisfied if the center of the banana is stably positioned on the plate.

The Block Stacking task is performed in the same workspace. The objects manipulated are $40\text{ mm}$ cubic blocks. The objective is to pick up an orange block from the right side of the workspace and stack it on top of a gray block on the left side. The task is considered successful if the orange block remains stably stacked on top without falling.

\subsection{Imitation Learning Hyperparameters}

\begin{figure}[htbp]
    \begin{minipage}[c]{0.52\textwidth}
        \centering
        \includegraphics[width=\linewidth]{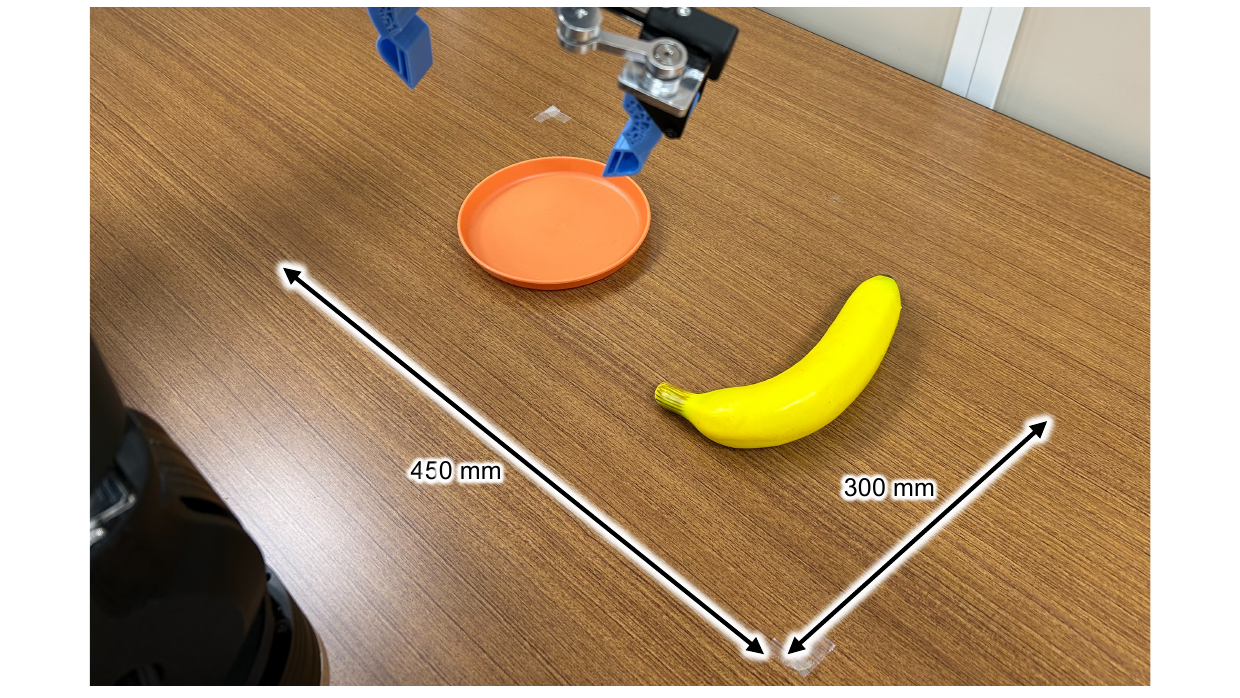}
        \caption{Experimental setup and workspace dimensions for the imitation learning tasks.}
        \label{fig:exsetup}
    \end{minipage}\hfill
    \begin{minipage}[c]{0.45\textwidth}
        \centering
        \captionof{table}{Hyperparameters for ACT Policy Training}
        \label{tab:act_params}
        \small
        \begin{tabular}{@{}lc@{}}
        \toprule
        \textbf{Hyperparameter} & \textbf{Value} \\ \midrule
        Learning Rate & $1 \times 10^{-5}$ \\
        Batch Size & 512 \\
        Action Chunk Size & 50 \\
        Hidden Dimension & 512 \\
        Feedforward Dimension & 3200 \\
        Encoder Layers & 4 \\
        Decoder Layers & 7 \\
        Attention Heads & 8 \\
        KL Weight ($\beta$) & 10 \\
        Training Epochs & 5000 \\ \bottomrule
        \end{tabular}
    \end{minipage}
\end{figure}

To ensure full reproducibility of the downstream imitation learning evaluation, Table \ref{tab:act_params} details the hyperparameters used for training the Action Chunking with Transformers (ACT) policies. We maintained identical hyperparameter configurations across all device conditions to ensure a fair comparison of the hardware's effect on data quality.

\end{document}